# On the Diagnostic of Road Pathway Visibility

**Pierre Charbonnier**
Research Dir.
LCPC (ERA 27)
11 rue Jean Mentelin, B.P. 9, 67035
Strasbourg, France
Pierre.Charbonnier@developpement-durable.gouv.fr

**Jean-Philippe Tarel**
Researcher
University Paris Est-INRETS-LCPC
(LEPSIS)
58 Bd Lefebvre, 75015, Paris, France
tarel@lcpc.fr

**François Goulette**
Associate professor
CAOR - Centre de Robotique - Mathématiques et Systèmes - Mines ParisTech
60 boulevard Saint Michel, 75272 Paris, France
francois.goulette@mines-paristech.fr

## Abstract

Visibility distance on the road pathway plays a significant role in road safety and in particular, has a clear impact on the choice of speed limits. Visibility distance is thus of importance for road engineers and authorities. While visibility distance criteria are routinely taken into account in road design, only a few systems exist for estimating it on existing road networks. Most existing systems comprise a target vehicle followed at a constant distance by an observer vehicle, which only allows to check if a given, fixed visibility distance is available. We propose two new approaches that allow estimating the maximum available visibility distance, involving only one vehicle and based on different sensor technologies, namely binocular stereovision and 3D range sensing (LIDAR). The first approach is based on the processing of two views taken by digital cameras onboard the diagnostic vehicle. The main stages of the process are: road segmentation, edge registration between the two views, road profile 3D reconstruction and finally, maximal road visibility distance estimation. The second approach involves the use of a Terrestrial LIDAR Mobile Mapping System. The triangulated 3D model of the road and its surroundings provided by the system is used to simulate targets at different distances, which allows estimating the maximum geometric visibility distance along the pathway. These approaches were developed in the context of the SARI-VIZIR PREDIT project. Both approaches are described, evaluated and compared. Their pros and cons with respect to vehicle following systems are also discussed.

## 1. Introduction

In this paper, we address the problem of assessing the visibility distance along an existing road path. Among the many definitions of visibility distances that can be found in the literature, we consider the stop-on-obstacle scenario. More specifically, we define the *required* visibility distance as the one needed by a driver to react to the presence of an obstacle on the roadway and to stop the vehicle. This distance clearly depends on factors such as the driver's reaction time or as the road's grip coefficient. They can be set to conventional, worst-case values e.g. 2 seconds for the reaction time. The grip coefficient is fixed to a value corresponding to a wet road surface.

The distance also depends, of course, on the speed of the vehicle, whose conventional value may be set to the so-called V85, i.e. to the 85$^{th}$ centile of the speed distribution. Naturally, speed statistics are not available at every point of the path. However, we can use the same laws as for road conception, stemming from statistical studies [1,2], to modulate a fixed conventional V85 value (that depends on the road classification) according to geometrical characteristics of the road, namely curvature and slope. The *required* visibility distance has to be compared to the *available* visibility distance, which is the maximal distance at which an object can be seen on the road as a function of the geometry of the road environment. We investigate two approaches for accurately estimating the available visibility distance.

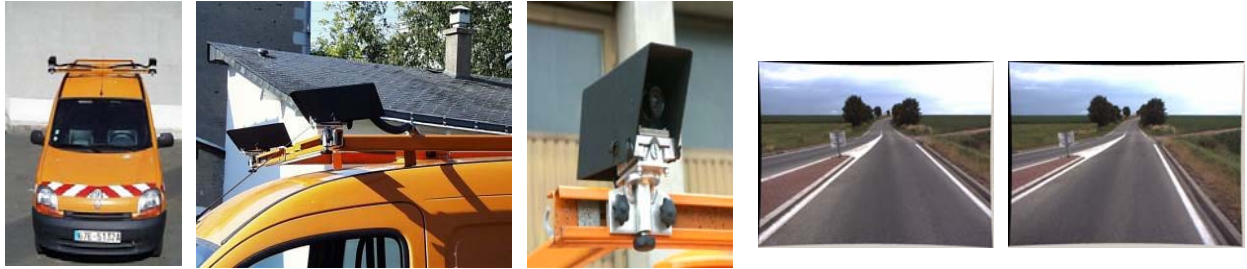

*Figure 1 : Experimental inspection vehicle with two digital cameras for stereovision (left 3 images, by courtesy of CECP). On the rightmost part of the figure, a stereo pair of images acquired by the vehicle is displayed.*

Since road visibility evaluation by drivers is essentially a visual task, the first approach we propose exploits images taken by cameras mounted on the inspection vehicle. The use of a single camera allows only 2D measures, and thus two cameras and stereovision are required to take into account the 3D shape of the road while estimating the visibility distance. The acquisition system being reduced to two color cameras and a computer, its is quite inexpensive. Pairs of views (see example on Fig. 1) can be acquired every 5 meters along the path, by the vehicle at a normal speed. However, the design of an accurate image processing method for estimating the 3D visibility distance remains challenging due to the difficulties of 3D reconstruction from pairs of real images taken under uncontrolled illumination in the traffic.

As an alternative, we also investigated the use of 3D data provided by a LIDAR terrestrial Mobile Mapping System, called LARA-3D. This prototype had been developed by the CAOR (Mines ParisTech) for several years [3]. Its adaptation to the creation of 3D road models for visibility computation has been initiated during the SARI-VIZIR project. While the vehicle moves, 3D points are sampled along the road and registered in an absolute reference system thanks to the use of INS and GPS sensors. In order to generate 3D models from points, several algorithms have been implemented, addressing particularly the need for downsizing the 3D model, and of artefacts suppression. For suppressing artefacts due to the presence of vehicles, we use the information of position of the vehicle on the road and approximate position of the borders, to suppress the data above it. For downsizing, we use at first a decimation step over each scanner profile, and then a plane modelling approach based on RANSAC and region growing, followed by a BPA triangulation [4]. These data processing lead to an important reduction of the number of triangles (factor 10 to more) and to a filtering of artefacts sufficient so that the 3D model can be used by the subsequent computation of visibility. This stage provides a model of the road environment as a triangulated 3D mesh, as shown in Fig. 2.

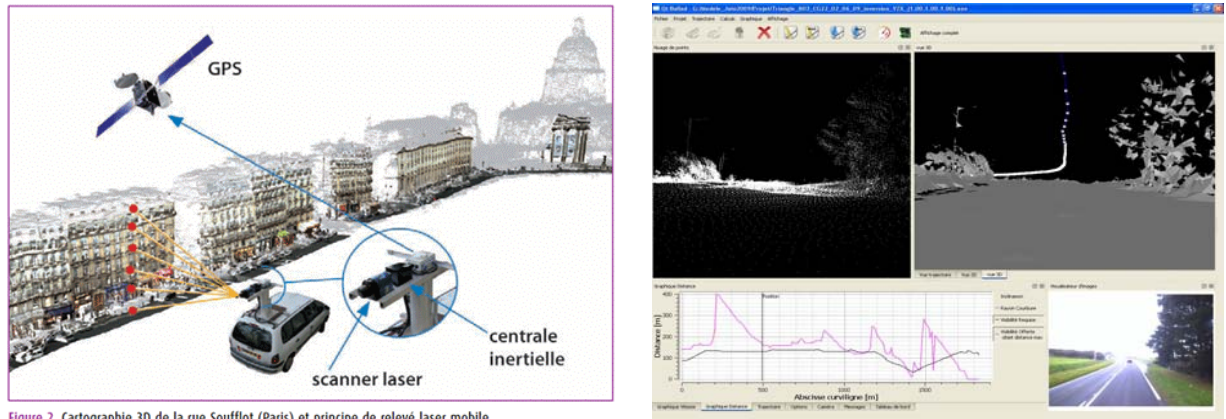

*Figure 2 : The LARA-3D experimental inspection vehicle of the CAOR (Mines ParisTech) is shown on the left with an example of obtained 3D point cloud and mesh on the right.*

Although the acquisition system is more complex – and hence, expensive - than in the stereovision approach, its advantage is that 3D data is directly grabbed with a high accuracy.

## 2. Visibility distance using binocular stereovision

In this case, the available visibility distance is defined as the maximum distance of points belonging to the image of the road. It is estimated by a three-step processing of the pair of left and right stereo images: segmentation of the roadway on each image using color information, registration between the edges in the road regions of the left and right images allowing to estimate a 3D model of the road surface, estimation of the maximum 3D distance of visible points on the road.

### 2.1 Road segmentation

The road being of main importance for estimating the visibility, the first step of the processing consists in classifying each pixel, in left and right images, as *road* or *non-road*. This segmentation processing is based on an iterative learning along the sequence of images of the road colors and of the non-road colors.

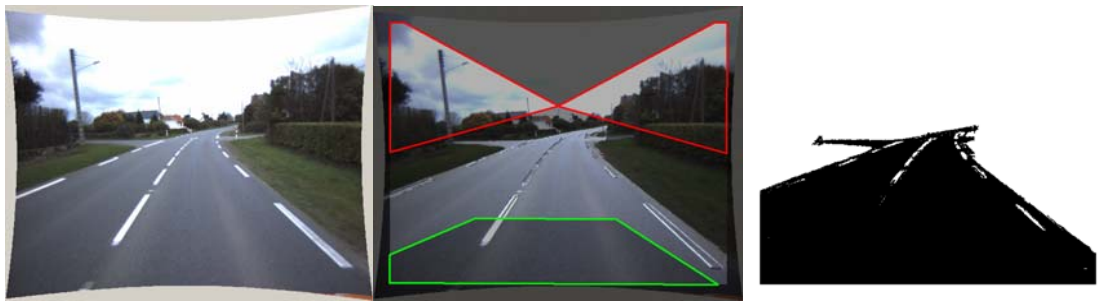

*Figure 3 : The original image of the front road on the left is processed learning road colors from pixel colors of the green region, and learning non-road colors from the red ones. The result of the classification of the pixels in road and non-road pixels is shown on the right.*

To this end, the pixels in the bottom center of each image are assumed with road colors, the pixels in the left and right top of the images are assumed with non-road colors, see the middle image in Fig. 3. The colors of road and non-road models are collected in past images when the processing is running real time, or better, collected using the following images in the sequence when the processing is in batch. Batch processing must be preferred since it allows sampling road colors at different distances ahead of the current image and thus leads to improved segmentation in the presence of lightening perturbations, shadows, variations of pavement color. Once the road and non-road color models are built, the image is segmented in two classes using a growing process of the road region starting from the bottom center part of the image, see Fig. 3 for an example of resulting segmentation. More details on this algorithm can be found in [5].

## 2.2 Edge registration and road profile reconstruction

The edges found in the left and right road regions segmented in the first step differ in pose due to the change in viewpoint. This difference in pose is directly related to the shape of the road surface.

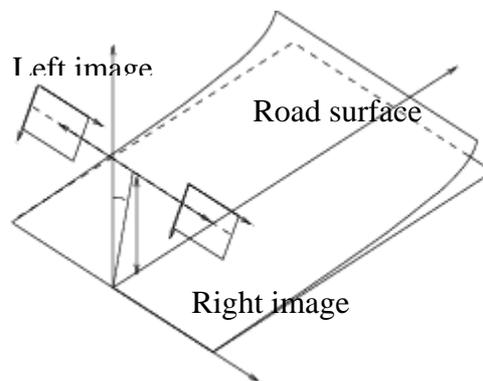

*Figure 4 : After matching the road edges in the left and right images, points on the road can be estimated by triangulation from the left and right viewpoints.*

Assuming a polynomial parametric model of the road longitudinal profile, see Fig. 4, the second step consists in the global registration of the left and right road edges and in the estimation of the parameters of the polynomial road model. This algorithm is described in details in [6] and is based on an iterative scheme that alternates between pairs of edge pixels registration and road surface parameters estimation.

## 2.3 Road visibility distance estimation

Once the 3D surface of the road has been estimated from the second step of the processing, the maximum distance the left and right registered edges is computed and is used as an estimate of the road visibility distance. Examples of results are shown in Fig. 5 where the red line corresponds to the visibility height in each image. As explained in details in [7], to qualify this estimated value of the visibility distance, the standard deviation of the estimator is also computed.

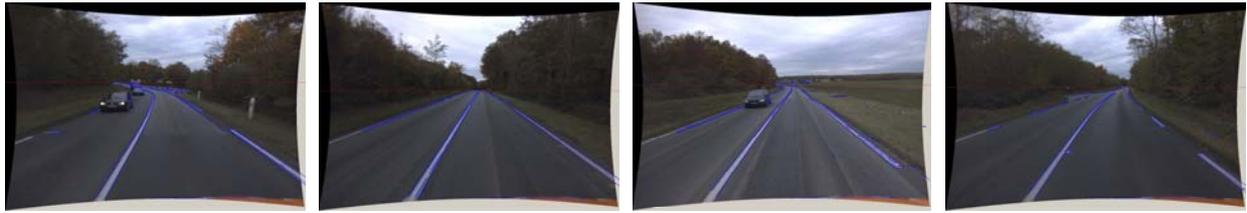

*Figure 5 : The blue edges are the edges of the right image registered on the left image, which is displayed. These images are extracted from a sequence of 400 images from road RD372, CG 91.*

## 3. Visibility distance from a 3D model of the road and its surroundings

In this approach, we exploit both the 3D triangulated data provided by the LIDAR system and the trajectory recorded by GPS/INS integration to estimate visibility distances.

### 3.1 Estimating the visibility distance from a 3D model

The definition of the available visibility distance we use in this approach is purely geometric and does not involve any photometric or meteorological consideration. It involves a target and an observer. The target may either be placed at a fixed distance from the observation point, to assess the availability of a certain prescribed distance, or moved away from it until it becomes invisible, to estimate the maximal visibility distance.

Conventional values can be found in the road design literature [1,2] for both the location of the observer and the geometry and position of the target. Both the viewpoint and the target are centered on the road lane axis. Typically, the viewpoint is located 1 meter high, which roughly corresponds to a mean driver's eye position. The conventional target is a pair of points that model a vehicle's rear lamps. The target is considered as visible as long as at least one point is visible, i.e. as long as it is possible to draw a straight line between the viewpoint and one of the target point without encountering any obstacle. Clearly, *ray-tracing* algorithms seem well-suited for this task. We also found it realistic to consider a parallelepiped (1.5×4×1.3 m) to model a vehicle. We found that a good rule of thumb was to consider the target as visible if 5% of its surface is visible. The threshold was set experimentally, in such a way that the results of the test would be comparable to the ones obtained when using the conventional target.

### 3.2 Qt-Ballad: a tool for visibility estimation

We have developed a specific software, called Qt-Ballad (see Fig. 6) during the SARI-VIZIR PREDIT project. It allows navigating in the 3D model (that can be visualized as well as a point cloud or a volumetric-rendered mesh). The trajectory of the road may be visualized in the 3D model. Volumetric or point-wise targets may be placed at different distances from the viewpoint. To make the interpretation easier, road scene images may be also visualized. All views are synchronised and the interface is completely reconfigurable.

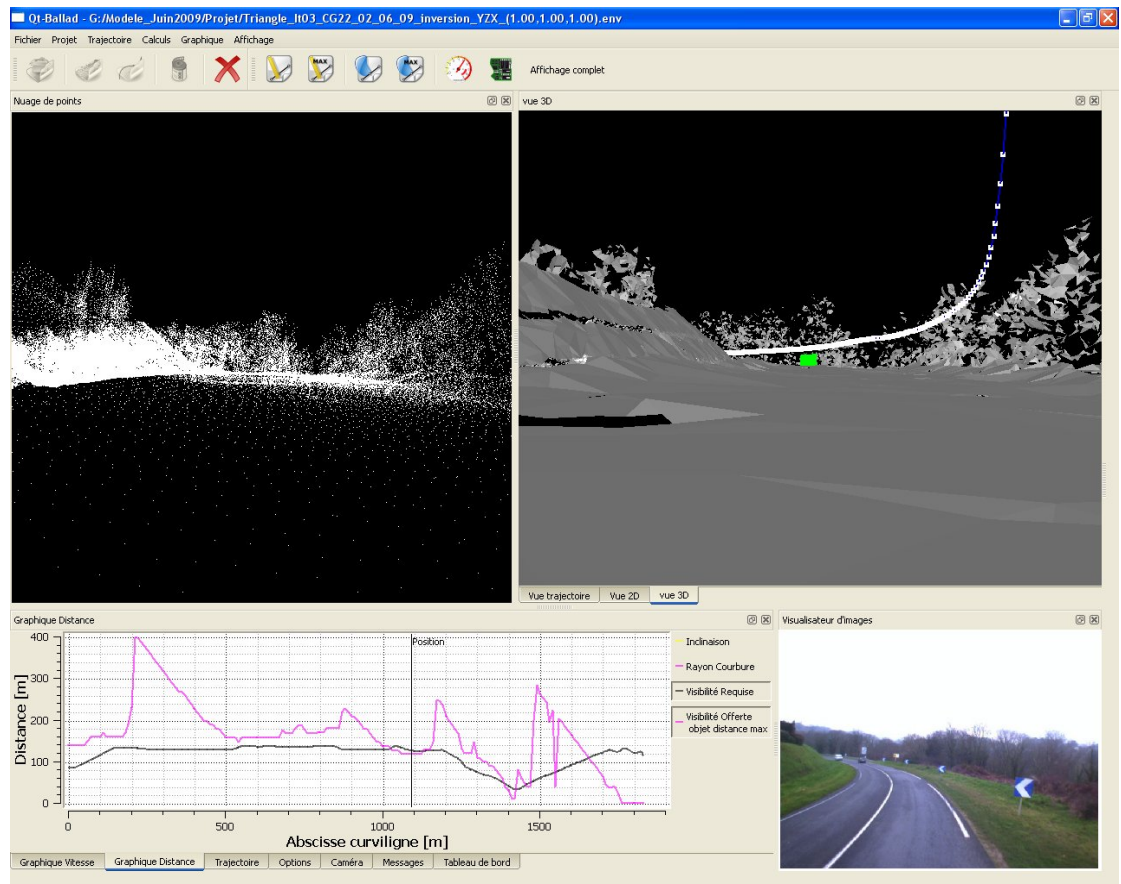

*Figure 6 : interface of the Qt-Ballad software. The top-left panel displays the original point cloud. The top-right one shows the triangulated model (in grey), the trajectory of the LIDAR system (in blue) and a parallelepiped target placed 100 m away from the observation point (in green). At the bottom right is displayed an image of the scene. Finally, the bottom-left panel displays the curve of required visibility (black) and of the estimated available visibility (magenta) vs the curvilinear abscissa. The vertical line shows the current position. The curves show that the available visibility is not sufficient in this situation.*

Qt-Ballad implements both required and available visibility distance computation. For the point-wise, conventional target, we use a software ray-tracing algorithm. When the volumetric target is used, we exploit the Graphical Processing Unit's capabilities. More specifically, the target is first drawn in the graphical memory, then the scene is rendered using Z-buffering and finally, an *occlusion-query* request (which is standard in up-to-date OpenGL implementations) provides the percentage of visible target surface. The visibility distance can be computed at every point of the trajectory (i.e. every 1 meter) or with a fixed step (typically, 5 or 10 meters) to speed up computations.

Two different ways of computing the visibility are implemented. In the first case, a fixed distance is maintained between the observation point and the target. The output is a binary function which indicates at every position whether the prescribed distance is available or not. In the second case, for every position of the observation point, the target is moved away as long as it is visible and the maximum available visibility distance is recorded.

## 4. Discussion

Within the SARI-VIZIR PREDIT project, see [8], different kinds of road section have been selected to perform a comparative study. Three approaches were tested: using stereovision, using a 3D model and using vehicle following. The last method, called VISULINE [9] is used as a reference. It consists of two vehicles connected by radio, that stay at a given fixed distance. An operator, in the second vehicle, checks visually whether the first vehicle is visible or not. To estimate the maximum visibility distance, the vehicle-following system had to be run for several intervals: 50, 65, 85, 105, 130, 160, 200, 250 and 280 meters. The result is shown on the 2 km road section RD768, CG22, as a green discontinuous curve, see Fig. 7. On the same figure, the required visibility distance computed from the road characteristics and shown as a red, continuous curve. The estimates of available visibility obtained using stereovision are displayed as a blue curve with pluses, and the black curve with stars show the results of the approach based on 3D modelling.

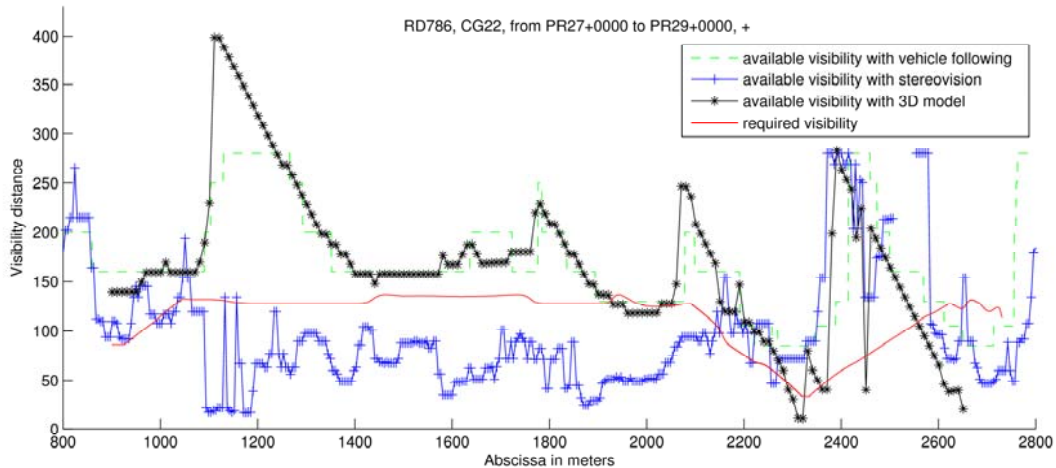

*Figure 7 : Comparison on the same 2 km road section (RD786, CG22), of the estimates of the available visibility distance using 3 techniques: vehicle following, stereovision, 3D model. The required visibility computed from the road geometry is also displayed. When the available visibility is lower that the required one (e.g. at abscissa 2000 m), a lack of visibility is detected.*

The results show that the stereovision approach under-estimates the visibility distance during the major part of the section. A careful examination of the results shown that the algorithm is very sensitive to the presence of vehicles in the road scene. Moreover, the long range road/non-road classification becomes more difficult when colors are flatter, which occurs under certain weather conditions or during the winter. The results of the 3D-based method are in good accordance with those of the reference method. However, we recall that the vehicle-following system requires several runs to obtain this result. Moreover, it is limited to 280 m while the maximal available distance seems to reach 400 m about abscissa 1150 m.

As a conclusion, this experiment shown that, as soon as an accurate 3D model of the road and its surrounding is available, well-known visualization algorithms such as *ray-tracing* or *z-buffering* suffice to estimate the available visibility. However, obtaining such models requires costly technologies and care must be taken during the processing of the point cloud. In particular, the triangulated model must be simplified enough to allow processing long road sections, but without losing too much useful information. On the other hand, the method based on image processing tends to underestimate the visibility distance due to the difficulties inherent to the task of segmenting the road far away from the camera. However, stereovision is cheap and can be very accurate in the near field. We therefore believe that it might provide an interesting alternative to LIDAR sensors and we plan to investigate this in the near future.

## 5. Acknowledgements

The authors are thankful to the French PREDIT SARI-VIZIR project for funding. The VISULINE data was provided by the LRPC St Brieuc (ERA 33 LCPC).